\documentclass[sn-mathphys,Numbered]{sn-jnl}

\usepackage{graphicx}%
\usepackage{multirow}%
\usepackage{amsmath,amssymb,amsfonts}%
\usepackage{amsthm}%
\usepackage{mathrsfs}%
\usepackage[title]{appendix}%
\usepackage{xcolor}%
\usepackage{textcomp}%
\usepackage{manyfoot}%
\usepackage{booktabs}%
\usepackage{algorithm}%
\usepackage{algorithmicx}%
\usepackage{algpseudocode}%
\usepackage{listings}%
\usepackage{array}

\usepackage{standalone}

\usepackage{scalerel}
\usepackage{tikz}
\usetikzlibrary{calc, shapes, arrows, positioning, patterns, decorations.pathreplacing, backgrounds, petri, automata, shadows, external, trees}
\usepackage{pgf}
\usepackage{pgfplots}
\usepackage{svg}

\usepackage{cleveref}

\usepackage{url}

\usepackage{subcaption}

\usepackage{gincltex}

\usepackage{enumitem}
\usepackage{mathtools}

\usepackage{siunitx}
\usepackage{stackrel}
\usepackage{accents}

\usepackage{tabularx}

\usepackage{todonotes}

\newcommand{\universe}[0]{\mathcal{U}}
\newcommand{\actuniverse}[0]{\mathcal{A}} 
\newcommand{\lpmuniverse}[0]{\universe_{\lpmsetvar}} 
\newcommand{\clustersetuniverse}[0]{\universe_{\partition}}
\newcommand{\eventuniverse}[0]{\universe_{ev}}

\newcommand{\rankingfuncuniverse}[0]{\universe_{\textit{rank}}}
\newcommand{\distancefuncuniverse}[0]{\universe_{\distfunc}}
\newcommand{\attributesuniverse}[0]{\universe_{attr}}
\newcommand{\attributevaluesuniverse}[0]{\universe_{val}}

\newcommand{\set}[1]{\{ #1 \}} 
\newcommand{\multiset}[1]{[\hspace{0.1em} #1 \hspace{0.1em}]} 
\newcommand{\seq}[1]{\langle #1 \rangle} 

\newcommand{\powerset}[1]{\mathcal{P}(#1)}
\newcommand{\allmultiset}[1]{\mathbb{M}(#1)} 
\newcommand{\allsequence}[1]{#1^*}
\newcommand{\proj}[2]{#1_{\upharpoonright_{#2}}} 
\newcommand{\partmapsto}[0]{\not\to} 
\newcommand{\st}[0]{\mid} 
\newcommand{\length}[1]{\vert #1 \vert} 
\newcommand{\norm}[1]{\lVert #1 \rVert} 
\newcommand{\seqvar}[0]{\sigma}

\newcommand{\concat}[0]{\cdot} 

\newcommand{\normalsub}[0]{\sqsubseteq} 
\newcommand{\relaxsub}[0]{\mathrel{\raisebox{1pt}{$\undertilde{\sqsubset}$}}} 
\newcommand{\overseq}[2]{\overline{#1}^{#2}}

\newcommand{\emarking}[0]{\multiset{}} 
\newcommand{\imarking}[0]{M_i}
\newcommand{\fmarking}[0]{M_f}

\newcommand{\lang}[1]{\mathcal{L}(#1)} 
\newcommand{\nlang}[2]{\mathcal{L}^{#2}(#1)}
\newcommand{\preset}[1]{\bullet #1} 
\newcommand{\postset}[1]{#1 \bullet} 

\newcommand{\fire}[1]{\xrightarrow{#1}} 
\newcommand{\fireseq}[0]{\sigma} 
\newcommand{\allfireseq}[2][]{\mathcal{F}_{#1}(#2)} 

\newcommand{\eventvar}{e}
\newcommand{\tracevar}[0]{\rho}
\newcommand{\eventlogvar}[0]{L}
\newcommand{\eventlogeventsvar}{E}
\newcommand{\eventseqvar}[0]{s}
\newcommand{\eventmapping}{\pi}
\newcommand{\eventattributevar}{attr}
\newcommand{\eventattributes}{Attr}
\newcommand{\eventattributesvalues}{\mathit{Val}}
\newcommand{\geteventattr}[2]{\eventmapping(#1, #2)}

\newcommand{\lpmsetvar}[0]{\textit{LPM}}
\newcommand{\lpmvar}[0]{\textit{lpm}}
\newcommand{\lpmnet}[1]{N_{#1}}
\newcommand{\allfireseqlpm}[1]{\allfireseq[\textrm{LPM}]{#1}}
\newcommand{\modelA}[0]{\lpmvar_A}
\newcommand{\modelB}[0]{\lpmvar_B}
\newcommand{\rank}[0]{\textit{rank}}
\newcommand{\freetransitions}[0]{T_{in}}
\newcommand{\lpmocurrencelist}[0]{\lambda} 

\newcommand{\simm}[1]{\mathit{sim}_{#1}}
\newcommand{\partition}[0]{\sqcap}

\newcommand{\collectivecontext}[0]{\vec{cc}}
\newcommand{\distfunc}[0]{\mathit{dist}}

\newcommand{\constname}[1]{\textit{#1}}
\newcommand{\lpmtext}[0]{LPM}

\DeclareMathOperator*{\argmax}{arg\,max}

\newcolumntype{M}[1]{>{\centering\arraybackslash}m{#1}}

\newcolumntype{C}{>{\centering\arraybackslash}X}

\newtheorem{definition}{Definition}%

\pgfplotsset{compat=1.18}

\begin{document}

{%
\thispagestyle{empty}
\vspace*{\fill}
\begin{center}
\fbox{%
\parbox{0.85\textwidth}{%
\centering\bigskip

{\large\textbf{Postprint Notice}}\bigskip

This version of the article has been accepted for publication, after peer review but is not the
Version of Record and does not reflect post-acceptance improvements, or any corrections.

\bigskip
The Version of Record is available online at:

Peeva, V., van der Aalst, W.M.P. (2026).
Framework for Grouping Local Process Models.
\textit{Journal of Intelligent Information Systems}, \textbf{64}(3), 941--964.\\[0.5em]
\url{https://doi.org/10.1007/s10844-025-01010-x}

\bigskip
This article is licensed under a Creative Commons Attribution 4.0 International Licence
(\url{http://creativecommons.org/licenses/by/4.0/}), which permits use, duplication, adaptation, distribution and
reproduction in any medium or format, as long as appropriate credit is given to the original author(s) and the source,
a link is provided to the Creative Commons licence, and any changes made are indicated.

\bigskip
}}
\end{center}
\vspace*{\fill}
\clearpage
}

    \title[Framework for Grouping Local Process Models]{Framework for Grouping Local Process Models}


    \author*[1]{\fnm{Viki} \sur{Peeva}}\email{peeva@pads.rwth-aachen.de}

    \author[1]{\fnm{Wil M.P.} \sur{van der Aalst}}\email{wvdaalst@pads.rwth-aachen.de}

    \affil[1]{\orgdiv{Chair of Process and Data Science (PADS)}, \orgname{RWTH Aachen University}, \orgaddress{\street{Ahronstrasse 55}, \city{Aachen}, \postcode{
        52074}, \state{North Rhine-Westphalia}, \country{Germany}}}

%


    \abstract{Local Process Models (\lpmtext{}s) are an underexplored concept in process mining.
    \lpmtext{}s describe patterns in event data considering sequence, choice, concurrency, and loop.
    In recent years, process mining has proved successful in the analysis and improvement of operational processes.
    More often than not, surprising findings are found when one does not consider the full process, making \lpmtext{}s
    and their discovery highly valuable.
    However, similar to other pattern mining approaches, \lpmtext{} discovery algorithms face the problems of model
    explosion and model repetition, i.e., the algorithms may create hundreds if not thousands of \lpmtext{}s, and
    subsets of them are close in structure or behavior.
    Practically, no analyst would be able to comb through thousands of \lpmtext{}s leading to using a sample of \lpmtext{}s
    that are easily accessible.
    The current sentiment is that the top-scoring \lpmtext{}s form the optimal sample to be presented.
    However, different applications should demand a different optimal sample.
    With this work, we show that if the goal of the mined \lpmtext{}s is to understand a process, using the top-scoring
    \lpmtext{}s as an optimal sample is a poor choice because of high repetition.
    We propose a framework for grouping \lpmtext{}s and creating an optimal sample by taking one representative \lpmtext{}
    for each group.
    We measure similarity between models via established process model similarity measures or by comparing the
    context in which an \lpmtext{} appears.
    The context is formed using data attributes available in the underlying event logs.
    We demonstrate the usefulness of grouping on multiple event logs by comparing repetition and coverage
    between samples comprised of the top-scoring models and the representatives of discovered groups.}




    \maketitle

    \section{Introduction}\label{sec:introduction}

    Process mining is a scientific discipline for discovering, monitoring, and improving processes via
    readily-available data from different data management systems.
    The three main pillars of process mining are process discovery, conformance checking, and process enhancement~\cite{DBLP:books/sp/Aalst16}.
    As the interest in process mining grows, the spectrum of processes to be analyzed expands.
    This triggers the need for novel techniques that provide insights into highly unstructured processes.
    Especially difficult, when analyzing such processes, is discovering a single well-defined model.
    Traditional process discovery techniques by trying to model the full range of variability would result in a
    so-called flower models (models that allow any behavior) or spaghetti models (models that because of modeling
    all the particularities in the process are difficult to read)~\cite*[See Fig. 7]{DBLP:journals/jides/TaxSHA16}.
    To resolve this problem, one usually focuses on frequent behavior~\cite{DBLP:conf/data/Aalst20}.
    However, in some domains like healthcare and IoT~\cite{DBLP:journals/jbi/Munoz-GamaMFJSH22,DBLP:journals/kais/BrzychczyAJK25},
    the rule does not hold, and better solutions are needed.

    \begin{figure}[t]
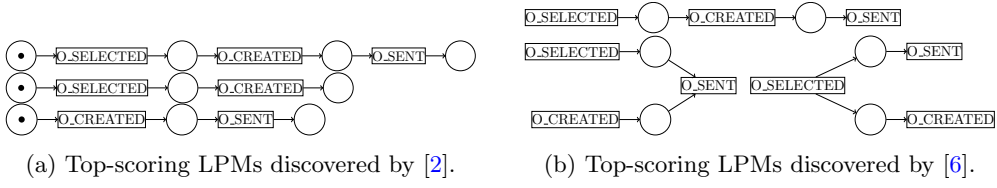

        \centering
        \begin{subfigure}[b]{0.475\textwidth}
            \includegraphics[width=\linewidth]{tax_lpm1.tex}
            \\
            \includegraphics[width=0.74\linewidth]{tax_lpm2.tex}
            \\
            \includegraphics[width=0.68\linewidth]{tax_lpm3.tex}
            \caption{Top-scoring \lpmtext{}s discovered by \cite{DBLP:journals/jides/TaxSHA16}.}
        \end{subfigure}
        \hfill
        \begin{subfigure}[b]{0.475\textwidth}
            \includegraphics[width=0.8\linewidth]{peeva_lpm1.tex}
            \\
            \includegraphics[width=0.45\linewidth]{peeva_lpm2.tex}
            \hfill
            \includegraphics[width=0.52\linewidth]{peeva_lpm3.tex}
            \caption{Top-scoring \lpmtext{}s discovered by \cite{DBLP:conf/apn/PeevaMA22}.}
            \label{fig:similar_models_peeva}
        \end{subfigure}
        \caption{\lpmtext{}s discovered using \cite{DBLP:journals/jides/TaxSHA16} and \cite{DBLP:conf/apn/PeevaMA22}
        on the filtered and transformed \textit{BPIC2012} event log for resource $10939$ as explained
        in~\cite{DBLP:journals/jides/TaxSHA16}.}
        \label{fig:similar_models}
    \end{figure}

    A relatively new field is Local Process Model (\lpmtext{}) discovery~\cite{DBLP:journals/jides/TaxSHA16}, where
    the idea is to build smaller process models explaining fragments of the behavior instead of one overall model.
    Yet, current \lpmtext{} discovery approaches return hundreds or thousands of models (i.e., \lpmtext{}s) for one
    event log (\emph{model explosion}), with highly similar models repeating between them (\emph{model repetition})
    as shown in \Cref{fig:similar_models}.
    Ideally, depending on the application in focus, one would consider a sample of
    the mined \lpmtext{}s, which we refer to as \emph{application-optimal} or \emph{optimal} if the application is
    clear or irrelevant.
    Current \lpmtext{} discovery approaches have introduced evaluation measures and rank the \lpmtext{}s based on
    these.
    With this, they implicitly establish the optimal sample to be the top-scoring \lpmtext{}s.
    However, here we argue that a better optimal sample can be chosen for the task of capturing the underlined process.

    In this work, we propose grouping the discovered \lpmtext{}s and for each group choosing one representative \lpmtext{}.
    The set of representative \lpmtext{}s would then characterize the optimal sample.
    Considering different needs and interests, the framework allows for two different perspectives when grouping the \lpmtext{}s.
    On one hand, we allow for grouping \lpmtext{}s based on the control flow by using already established process model
    similarity measures.
    On the other hand, we group \lpmtext{}s based on the data attributes collected during the process execution.
    Meaning, \lpmtext{}s are considered similar when they cover events appearing in a similar context.
    In the evaluation, we focus on comparing the top-scoring samples to the group representative samples.
    We compute diversity between the models in a sample and the count of events a sample covers as measurable way to
    quantify how well a set of \lpmtext{}s capture the process.
    Additionally, we demonstrate the benefit of clustering \lpmtext{}s by inspecting a smaller set of models on the
    real-life event logs \constname{BPIC2012-res10939} and \constname{Sepsis}.
    This paper extends the work presented in~\cite{DBLP:conf/icpm/PeevaA23} by incorporating the data
    perspective  when grouping \lpmtext{}s.
    Furthermore, we introduce event coverage on set level to assess how much the selected sample fits the event log.
    Finally, we extend the experiments to incorporate additional clustering parameters, and new similarity notions.

    The reminder of the paper is structured as follows.
    In~\Cref{sec:related-work}, we describe other grouping methods, and we cover related work of sub-parts of our framework.
    With~\Cref{sec:preliminaries}, we establish the foundations necessary to follow the rest of the paper.
    In \Cref{sec:grouping-framework}, we describe all contributions in detail, and we present the obtained results
    in \Cref{sec:evaluation_results}.
    Finally, we discuss the limitations of the approach and future work in \Cref{sec:discussion} and conclude the
    paper with \Cref{sec:conclusion}.


    \section{Related Work}\label{sec:related-work}
    Our framework for grouping similar \lpmtext{}s together is based on multiple key aspects.
    Here, we consider related work for those individually.
    First, we consider \lpmtext{} discovery approaches and how they incorporate data attributes in~\Cref{subsec:rw-local-process-models}.
    Then, we discuss existing process model similarity measures and other frameworks for grouping \lpmtext{}s or
    process models in~\Cref{subsec:rw-grouping-process-models}.

    \subsection{Local Process Models}\label{subsec:rw-local-process-models}

    \lpmtext{} discovery is positioned in-between traditional process discovery and episode and sequence mining~\cite{DBLP:journals/datamine/MannilaTV97,DBLP:conf/simpda/LeemansA14a,DBLP:conf/icde/AgrawalS95,DBLP:conf/edbt/SrikantA96}.
    In contrast to process discovery~\cite{DBLP:conf/bpm/BergenthumDLM07,DBLP:conf/bpm/CarmonaCK08,DBLP:conf/apn/LeemansFA13,DBLP:conf/apn/MannelA19,DBLP:journals/datamine/WenAWS07,DBLP:journals/fuin/derWerfDHS09,DBLP:journals/computing/ZelstDAV18}, whose task is to discover one model that explains the traces of an event log from start to end, \lpmtext{} discovery tries to mine a set of models each matching some particular behavioral patterns represented by subsequences in the event log.
    While episode and sequence mining can also be used to find frequent patterns in event logs, \lpmtext{}s allow for more complex constructs like choices and loops.
    They were first introduced in~\cite{DBLP:journals/jides/TaxSHA16} as a replacement for process discovery of highly
    unstructured processes.
    However, afterward, the use cases of \lpmtext{}s expanded (see~\cite{DBLP:conf/bpm/DeevaW18,DBLP:conf/sac/DelcoucqLFA20,DBLP:journals/widm/GuzzoRV22,DBLP:conf/ewgdss/KirchnerM18,DBLP:conf/otm/LeemansTH18,DBLP:journals/is/MannhardtLRAT18,DBLP:conf/emisa/MannhardtT17,PijnenborgVFLG21}) and multiple approaches and extensions for \lpmtext{} discovery followed~\cite{DBLP:journals/topnoc/DalmasTN18,DBLP:conf/caise/AcheliGW19,DBLP:journals/tkde/AcheliGW22,DBLP:conf/apn/PeevaMA22,DBLP:conf/icpm/PeevaA23,DBLP:conf/icpm/BruningsF022}.

    The initial approach by~\citeauthor{DBLP:journals/jides/TaxSHA16} recursively extends process trees by appending more activities using different operators.
    In each iteration, because of monotonicity properties, \lpmtext{}s are filtered by frequency, and only those satisfying a threshold proceed to the next phase to be extended with additional activities.
    The algorithm stops when there are no more frequent \lpmtext{}s to be extended or new activities to be added.
    In~\cite{DBLP:conf/caise/AcheliGW19} two existing \lpmtext{}s (represented as process trees) differing only in one node are joined together to create a larger \lpmtext{}.
    In the new model, the differing nodes are replaced with a process tree operator and are added to it as children.
    Similarly, using monotonicity properties, at each step candidate \lpmtext{}s are pruned out.
    The approach in~\cite{DBLP:conf/icpm/BruningsF022} first finds frequently occurring gaped sequences and then discovers models on them.
    Finally, in~\cite{DBLP:conf/apn/PeevaMA22}, \lpmtext{}s are created by joining single place nets into larger models.
    All approaches can represent the same or highly similar behavior with multiple, nearly identical, models.
    Therefore, focusing on frequent behavior results in clusters of repetitive models among the highly-ranked models.

    All the approaches above~\cite{DBLP:conf/caise/AcheliGW19,DBLP:conf/apn/PeevaMA22,DBLP:journals/jides/TaxSHA16}, except for~\cite{DBLP:conf/icpm/BruningsF022}, suffer from the \emph{model explosion} and \emph{repetition} problems.
    In~\cite{DBLP:conf/icpm/BruningsF022}, the explosion problem is solved because for each frequent gapped sequences only one model is mined.
    However, this restricts the set of application scenarios the approach can be applied to (e.g., not suitable for
    event abstraction, trace classification and clustering, etc.), and even in the fewer
    built models, the repetition problem is still clearly visible.
    Moreover, for the most of the approaches there exist different extensions.
    For example, in~\cite{DBLP:conf/ssci/TaxSAH16} the authors try to decrease the complexity of~\cite{DBLP:journals/jides/TaxSHA16} by using heuristics to project the event log on subset of activities before mining local process models.
    Both~\cite{DBLP:journals/is/TaxDSAN18} and~\cite{DBLP:journals/tkde/AcheliGW22} use data attributes to restrict the context for which \lpmtext{}s are mined.
    Additionally, in~\cite{DBLP:journals/is/TaxDSAN18}, it is possible to define utility functions on top of data attributes.
    One can then mine \lpmtext{}s given a specific interest, by giving preference, i.e., higher rank, to \lpmtext{}s that score well on the utility functions.

    \subsection{Grouping Process Models}\label{subsec:rw-grouping-process-models}
    This paper tries to decrease the repetition of similar \lpmtext{}s by grouping them together.
    However, the challenge of organizing process models in model repositories~\cite{DBLP:journals/infsof/YanDG12},
    and querying~\cite{DBLP:journals/cii/Jin0RHW13,DBLP:conf/edoc/AwadPW08},
    grouping~\cite{DBLP:journals/eswa/OrdonezOCLWT17,DBLP:conf/bpm/AiolliBS11}, or
    merging~\cite{DBLP:journals/tosem/RosaDUD13,DBLP:journals/dss/SunKY06} similar process models is not new.
    For example,~\cite{DBLP:conf/bpm/QiaoAR11,DBLP:conf/ic3ina/SarnoGPS13} group process models by performing
    graph-based partitioning such that two process models are connected in the graph if the computed similarity
    between the two process models exceeds a predefined threshold. \cite{DBLP:journals/eswa/OrdonezOCLWT17} also
    presents an incremental clustering based on graph theory, but they do not prefilter connections between process
    models with low similarity, and the precomputed similarity matrix is directly used by the algorithm.
    In~\cite{DBLP:conf/iccsa/JungB06} process models are grouped using a hierarchical clustering algorithm, by
    converting them to activity transition vectors, and using the cosine measure to compute similarity.
    A hierarchical clustering method is also proposed in~\cite{su11092560}, but the way similarity between processes
    is computed is specific for manufacturing processes.
    In~\cite{DBLP:conf/iccsa/JungB06,DBLP:conf/ki/BergmannMW13} clustering is investigated for workflows,
    and~\cite{DBLP:conf/zeus/HeinzeAS21,DBLP:conf/bpm/EkanayakeDGRH12} consider Single-Entry Single-Exit (SESE)
    fragments in process models and try to discover approximate clones of such fragments in a process model repository.

    Most of the grouping approaches rely on some kind of a process model similarity measure.
    Along with grouping, operations like retrieval, model querying, and model addition without creating duplicates are
    challenging and need a way to measure process model similarity.
    Therefore, various similarity measures are proposed in the literature and one can categorize them in different ways.
    Over the years there have been multitude of survey papers~\cite{DBLP:books/daglib/p/DijkmanD13,DBLP:journals/debu/DumasGD09,DBLP:journals/csur/SchoknechtTFOL17,DBLP:conf/bpm/ThalerSFO016,DBLP:journals/cii/BeckerL12} covering this topic.
    Although there can be small differences in how survey papers categorize different similarity measures, all of
    them agree, the basic split is into measures that compare the structure of the process model, those that
    compare the behavior, or combine both.
    Subsequently, one can consider the level of abstraction used, e.g., complete language versus weak order relations.
    For more information, we direct the reader to the aforementioned survey papers.
    An alternative are similarity measures based on attributes.
    In~\cite{DBLP:journals/is/DijkmanDDKM11} a similarity measure that uses node attributes in a process model is proposed.
    However, this is different from our work, since we use data attributes available in the recorded behavior of the process.
    Such data attributes are used for clustering traces in~\cite{DBLP:conf/bpm/ZelstC20}, but we differentiate by
    using it to cluster models representing patterns in the underlying event log.

    When it comes to \lpmtext{}s, and the previously introduced \lpmtext{} discovery approaches, the
    implementations of~\cite{DBLP:journals/jides/TaxSHA16} and~\cite{DBLP:conf/apn/PeevaMA22} offer
    grouping of the models by considering the common transition labels.
    Additionally, in~\cite{DBLP:conf/icpm/PeevaA23} \lpmtext{}s are grouped by using process model similarity
    measures, but to the best of our knowledge, data attributes have not yet been used to group process models or \lpmtext{}s together.


    \section{Preliminaries}\label{sec:preliminaries}
    In this section, we provide essential background information needed to follow the rest of the paper.

    \subsection{Basic Definitions}\label{subsec:preliminiaries-general}
    We define sets ($X = \set{a,b}$), multisets ($ M = [a^2,b^3]$), sequences ($\seqvar = \seq{a,b,c}$), and tuples ($t
    = ( a,b,c )$) as usual.
    Given a set $X$, $\powerset{X}$ is the power set of $X$, $\allsequence{X}$ represents the set of all sequences over $
    X$, and $\allmultiset{X}$ is the set of all multisets over $X$.
    We use $\seqvar(i)$ to denote the $i$-th element of the sequence $\seqvar$, and $M(a) = 2$ to denote that item $a$
    appears twice in the multiset $M$.

    Given a sequence $\fireseq = \seq{\fireseq_1, \fireseq_2, \dots \fireseq_n}$, we extend $\fireseq$ with an
    additional element $\fireseq_{n+1}$ by writing ${\fireseq \cdot \fireseq_{n+1}}$.
    We call the sequence $\fireseq'$ a \emph{subsequence} of $\fireseq$, if and only if
    $\fireseq' = \seq{\fireseq_l, \fireseq_{l+1}, ... \fireseq_m}$ and ${1 \leq l \leq m \leq n}$ (we write
    $\fireseq' \normalsub \fireseq$).
    We call $\fireseq'$ a \emph{relaxed subsequence} (we write $\fireseq' \relaxsub \fireseq$) if and only if
    $\fireseq' = \seq{\fireseq_{i_1}, \fireseq_{i_2}, ... \fireseq_{i_k}}$ for
    $1 \leq i_1 < i_2 < \dots < i_k \leq n$ and $k \geq 1$, i.e., we drop any number of elements from $\fireseq$
    and keep the order for the rest.
    Lastly, given a sequence $\fireseq$ and a relaxed subsequence $\fireseq' \relaxsub \fireseq$, we call
    $s = \overseq{\fireseq'}{\fireseq}$ a \emph{minimal over-sequence} of $\fireseq'$ in $\fireseq$ if and only if
    $s \normalsub \fireseq \land \fireseq' \relaxsub s \land \nexists_{s'} (s' \normalsub \fireseq \land \fireseq' \relaxsub s' \land \length{s'} < \length{s})$.

    To recalculate sets or multisets from other sets, multisets, or sequences, we use the $\set{\cdot}$ and
    $\multiset{\cdot}$ operators.
    We write $f \colon X \to Y$ to denote functions, and $dom(f) = X$, $rng(f) = Y$ to
    denote the domain and the range of the function $f$ respectively.
    We use $f(X)=\{f(x) \st x \in X\}$ (and $f(\seqvar)=\langle f(\seqvar(1)), f(\seqvar(2)),\dots, f(\seqvar(n) \rangle$) to apply the function $f$ to every element in the set $X$ (the sequence $\seqvar$).
    Finally, we write $\proj{\seqvar}{X}$ to denote the projection of the sequence $\seqvar$ on the set $X$.

    \subsection{Process Mining}\label{subsec:preliminaries-process-mining}
    Normally, the collected data used for process analysis is transformed into \emph{event logs}.
    In traditional process mining, event logs focus on a single case notion where each event occurs exactly
    for one case.
    Events are denoted by the executed activity, the timestamp when the event occurred, and for which case they were
    executed.
    Additionally, event logs can contain further data attributes, either on case or event level.

    In \Cref{def:event-log}, we formally define \emph{event logs}, and with \Cref{def:traces} we extract the set of all
    traces of the event log, where a trace is defined as a sequence of events.

    \begin{table}[t]
        \centering
        \caption{A toy event log $\eventlogvar_{gp}$ for game play. The \constname{Event Id} column
        represent the events, and the rest of the columns represent different event attributes.}
        \begin{tabularx}{\textwidth}{|M{1cm}|M{1.2cm}|M{1.2cm}|C|C|M{2.1cm}|C|}
            \hline
            Event Id & Game Id (\textit{case}) & Activity (\textit{act}) &
            Timestamp (\textit{time}) & Player (\textit{player}) & Experience Tier (\textit{tier})
            & Idle Units (\textit{idle-units})
            \\ \hline
            $e1$ & 202 & \textbf{S}elect & 2024-02-13 16:57:21.030 & Trolldor & medium
            & 27
            \\ \hline
            $e2$ & 202 & \textbf{M}ove & 2024-02-13 16:57:23.047 & Trolldor & medium
            & 23
            \\ \hline
            $e3$ & 243 & \textbf{M}ove & 2024-02-13 16:57:24.025 & DustGod & advanced
            & 1
            \\ \hline
            $e4$ & 202 & \textbf{S}elect & 2024-02-13 16:57:28.015 & Trolldor & medium
            & 31
            \\ \hline
            $e5$ & 202 & \textbf{C}ollect & 2024-02-13 16:57:29.001 & Trolldor & medium
            & 16
            \\ \hline
            $e6$ & 202 & \textbf{A}ttack & 2024-02-13 16:58:03.066 & Trolldor & medium
            & 16
            \\ \hline
            $e7$ & 243 & \textbf{S}elect & 2024-02-13 16:58:04.011 & DustGod & advanced
            & 6
            \\ \hline
        \end{tabularx}
        \label{tab:example-event-log-gameplay}
    \end{table}

    \begin{definition}[Event Log]
        \label{def:event-log}
        Let $\eventuniverse$ be the universe of events, $\attributesuniverse$ the universe of attribute names,
        and $\attributevaluesuniverse$ the universe of attribute values.
        An \emph{event log} $\eventlogvar = (\eventlogeventsvar, \eventattributes, \eventmapping)$ is a tuple,
        with $E \subseteq \eventuniverse$ the set of events, $\eventattributes$ a set of attribute names
        such that $\set{act, case, time} \subseteq \eventattributes$, and $\eventmapping \colon \eventlogeventsvar
        \times \eventattributes \partmapsto \eventattributesvalues$ a (partial) function assigning to an event
        $\eventvar \in \eventlogeventsvar$ and attribute name $\eventattributevar \in \eventattributes$ pair, a value
        $\geteventattr{\eventvar}{\eventattributevar} \in \eventattributesvalues$. 
    \end{definition}

    \begin{definition}[Traces]
        \label{def:traces}
        Given an event log $\eventlogvar = (\eventlogeventsvar, \eventattributes, \eventmapping)$ and a case
        $c \in \geteventattr{\eventlogeventsvar}{case}$\footnote{We overload the function $\eventmapping$ on sets of
        events as explained in~\Cref{subsec:preliminiaries-general}.}, we define
        $\eventlogeventsvar^{\eventlogvar}_c = \set{\eventvar \in \eventlogeventsvar \st \geteventattr{\eventvar}{case} = c}$ as the set of all events of the case $c$.
        Now, the \emph{trace} of $c$ is a sequence of its events ordered by time, i.e., $\tracevar^{\eventlogvar}_c = \seq{e^c_1, e^c_2, ..., e^c_{\length{\eventlogeventsvar^{\eventlogvar}_c}}}$ where $\forall_{1 \leq i < j \leq \length{\eventlogeventsvar^{\eventlogvar}_c}} \geteventattr{e^c_i}{time} \leq \geteventattr{e^c_j}{time}$.
        Moreover, we write $traces(\eventlogvar) = \set{\tracevar^{\eventlogvar}_c \st c \in \geteventattr{\eventlogeventsvar}{case}}$ to denote all traces for the event log $\eventlogvar$.
    \end{definition}

    Let us consider the toy event log $\eventlogvar_{gp} = (\eventlogeventsvar, \eventattributes, \eventmapping)$
    simulating a Real Time Strategy (RTS) game in~\Cref{tab:example-event-log-gameplay}.
    In the event log, one case is considered one gameplay from the perspective of one user, meaning the attribute
    \constname{Game Id} represents the case.
    The \constname{Activity} and \constname{Timestamp} are the main attributes available for events, \constname{Player}
    and \constname{Experience Tier} are additional attributes on case level, and \constname{Idle Units} is an attribute
    on event level.
    Per \Cref{def:event-log}, $\geteventattr{e6}{case} = 202$, the $\geteventattr{e6}{player} = Trolldor$, and
    $\geteventattr{e6}{\textit{idle-units}} = 16$.
    The set of all events for the case $202$ is
    $\eventlogeventsvar^{\eventlogvar_{gp}}_{202} = \set{e1, e2, e4, e5, e6} \subseteq \eventlogeventsvar$,
    and the traces of the event log, we obtain with
    $traces(\eventlogvar_{gp}) = \set{\seq{e1, e2, e4, e5, e6}, \seq{e3, e7}}$.

    The behavior recorded in logs can be modeled using different notations, such as DFG, process trees, BPMN, Petri nets, etc.
    In this work, we focus on Petri nets, and more precisely on labeled Petri nets which we define in \Cref{def:petri-net}.
    Note that a transition $t \in T$ with $l(t) = \tau$ is
    called silent, and if two transitions share a common label, i.e., $t_1, t_2 \in T$ such that $t_1 \neq t_2$ and $l(t_1) = l(t_2)$, we call them duplicate transitions.

    \begin{definition}[Labeled Petri net]
        \label{def:petri-net}
        Given the universe of activities, $\actuniverse$, a \emph{labeled Petri net} $N = (P, T, F, l)$ is a tuple, where $P$ is a set of places and $T$ is a set of
        transitions such that $P \cap T = \emptyset$. $F \subseteq (P \times T) \cup (T \times P)$ is the flow relation,
        and $l: T \to \actuniverse \cup \set{\tau}$ the labeling function.
    \end{definition}

    Now, given a node $x \in P \cup T$, we define the \emph{preset} of $x$ as $\preset{x} = \set{y \in P \cup T
    \st (y, x) \in F}$ and the \emph{postset} of $x$ as $\postset{x} = \set{y \in P \cup T \st (x, y) \in F}$.

    To attach behavior to labeled Petri nets we use a \emph{marking} and the \emph{firing rule}.
    A marking $M$ denotes the state of a Petri net $N = (P, T, F, l)$ as a multiset of places ($M \in \allmultiset{P}$) and the firing rule allows for changes between states (i.e., markings).
    Given a marking $M$, a transition $t$ is \emph{enabled} in the marking $M$ if and
    only if $\preset{t} \subseteq M$.
    If a transition $t$ is enabled in the marking $M$, it can fire and change the state
    of the net to a new marking $M' = (M \setminus \preset{t}) \cup \postset{t}$.
    We write $M \fire{t} M'$.
    A sequence of transitions $\fireseq = \seq{t_1, \dots, t_n} \in \allsequence{T}$ is enabled in $M$ and by firing it
    marking $M'$ is reached if and only if there exist $M_0$, $M_1$, $\dots$, $M_n$ such that $M_0=M$, $M_n=M'$, and $M_{
        i-1}\fire{t_i}
    M_i$ for $1 \leq i \leq n$.
    We write $M \fire{\fireseq} M'$.

    Now, we formally define an \emph{accepting Petri net} in \Cref{def:accepting-petri-net} and the set of all its firing
    sequences in \Cref{def:accepting-petri-net-traces}.

    \begin{definition}[Accepting Labeled Petri Net]
        \label{def:accepting-petri-net}
        An \emph{accepting labeled Petri net} is a triple $(N, \imarking, \fmarking)$ such that $N = (P,T,F,l)$ is a
        labeled Petri net, $\imarking \in \allmultiset{P}$ is the initial marking, and $\fmarking \in \allmultiset{P}$ is
        the final marking.
    \end{definition}

    \begin{definition}[Complete Firing Sequences]
        \label{def:accepting-petri-net-traces}
        Let $\textrm{AN} = (N, \imarking, \fmarking)$ be an accepting Petri net, $\allfireseq{\textrm{AN}} = \set{\fireseq \in \allsequence{T} \st \imarking \fire{\fireseq} \fmarking}$ is the set of complete firing sequences for $\textrm{AN}$.
    \end{definition}

    The \emph{language} of an accepting Petri net $\textrm{AN} = (N, \imarking, \fmarking)$ is obtained by projecting all complete firing sequences on the transition labels and removing $\tau$-skips, i.e., $\lang{\textrm{AN}} = \set{\proj{l(\fireseq)}{\actuniverse} \st \fireseq \in \allfireseq{\textrm{AN}}}$.
    The sequences in $\lang{\textrm{AN}}$ we also call \emph{traces} of the net.
    We use $\nlang{\textrm{AN}}{n} = \set{\proj{l(\fireseq)}{\actuniverse} \st \fireseq \in \allfireseq{\textrm{AN}} \land \length{\fireseq} \leq n}$ to denote the language restricting to complete firing sequences of length at most $n$.

    \subsection{Local Process Models}\label{subsec:preliminaries-lpms}
    We now focus on formalizing \lpmtext{}s. Similar to end-to-end process models, \lpmtext{}s can be represented by
    any modeling language (Petri nets, process trees, BPMNs, etc.), however, in this work we restrict to a subclass of accepting labeled Petri nets.

    \begin{figure}[t]
        \centering
        \includegraphics[width=0.45\linewidth]{example_lpm.tex}
        \caption{Example \lpmtext{} $\lpmvar_{gp}$ for the Game Play event log in~\Cref{tab:example-event-log-gameplay}.}
        \label{fig:example-lpm-gp-log}
    \end{figure}

    We show an example \lpmtext{} in~\Cref{fig:example-lpm-gp-log}, and we give a formal definition in \Cref{def:lpm}.

    \begin{definition}[Local Process Models]
        \label{def:lpm}
        A \emph{Local Process Model} (\lpmtext{}) is an accepting labeled Petri net $\lpmvar = (N_{\lpmvar}, \imarking, \fmarking)$ such that $N_{\lpmvar}=(P, T, F, l)$ is a labeled Petri net that satisfies the following restrictions:
        \begin{enumerate}
            \item $\forall_{x, x' \in P \cup T} \, \exists_{\seq{x_1, \dots, x_n}} (x = x_1 \land x' = x_n \land \forall_{1\leq i < n} ((x_i, x_{i+1}) \in F) \lor (x_{i+1}, x_{i}) \in F))$, i.e., there is only one connected component, and
            \item $\forall_{p \in P}(\preset{p} \neq \emptyset \land \postset{p} \neq \emptyset)$, i.e., each place has
            at least one incoming and one outgoing arc,
        \end{enumerate}
        and the initial and final marking are empty, i.e., $\imarking = \emarking$ and $\fmarking = \emarking$.
        We use $\lpmuniverse$ to denote the universe of such \lpmtext{}s.
    \end{definition}

    We use $\freetransitions(\lpmnet{\lpmvar}) = \set{t \in T_{\lpmvar} \st \preset{t} = \emptyset}$ to denote the transitions that have an empty preset, and we call them \emph{unrestricted transitions}.
    In the case of the \lpmtext{} $\lpmvar_{gp} = (\lpmnet{\lpmvar_{gp}}, \emarking, \emarking)$, $P_{\lpmvar_{gp}} = \set{p1, p2}$, $T_{\lpmvar_{gp}} = \set{S, G, M, C, A}$, $F_{\lpmvar_{gp}} = \set{(S, p1), (G, p1), (p1, M), (p1, C), (M, p2), (C, p2), (p2, A)}$, and $\freetransitions(\lpmvar_{gp}) = \set{S, G}$.

    We define the set of complete valid firing sequences of such \lpmtext{}s in \Cref{def:lpm-behavior} by restricting
    that each place in the net can receive at most one token from unrestricted transitions.
    Meaning, for the \lpmtext{} from before, we have $\allfireseqlpm{\lpmvar_{gp}} = \set{\seq{S, M, A}, \seq{G, M, A}, \seq{S, C, A}, \seq{S, M, A}}$.
    Note, the firing sequence $\seq{S, M, G, C, A}$ is not a valid firing sequence, because two unrestricted transitions, namely $S$ and $M$, put a token in $p1$.

    \begin{definition}[\lpmtext{} Behavior]
        \label{def:lpm-behavior}
        Given an \lpmtext{} $\lpmvar=(N_{\lpmvar}, \imarking, \fmarking)$ such that $N_{\lpmvar}=(P, T, F,l)$, we define $\allfireseqlpm{\lpmvar} = \set{\fireseq \in \allfireseq{\lpmvar} \st \forall_{1 \leq i < j \leq \length{\fireseq}} (\fireseq(i) \in \freetransitions \land \fireseq(j) \in \freetransitions \implies \postset{\fireseq(i)} \cap \postset{\fireseq(j)} = \emptyset) }$
        to be all valid complete firing sequences of $\lpmvar$.
    \end{definition}

    We obtain the \emph{language} $\lang{\lpmvar}$ of an \lpmtext{} $\lpmvar$ analogous to the language of
    accepting Petri nets, but now we only consider the valid complete firing sequences as per \Cref{def:lpm-behavior}.
    We use $\nlang{\lpmvar}{n}$ to denote the language restricting to valid complete firing sequences of length at most $n$.

    Given an \lpmtext{} $\lpmvar$ and an event log $\eventlogvar$, we can extract all occurrences of the \lpmtext{} in
    the event log.
    To achieve this, we define a function $\lpmocurrencelist(\lpmvar,\eventlogvar)$ in~\Cref{def:lpm-occurrence-list}.

    \begin{definition}[\lpmtext{} Occurrence List]
        \label{def:lpm-occurrence-list}
        Given an \lpmtext{} $\lpmvar$ and an event log $\eventlogvar$, we define $\lpmocurrencelist(\lpmvar,\eventlogvar) = \set{\eventseqvar \st \exists_{\tracevar \in traces(\eventlogvar)} (\eventseqvar \relaxsub \tracevar \land \geteventattr{\eventseqvar}{\textit{act}} \in \lang{\lpmvar})}$\footnote{We overload the function $\eventmapping$ on sequences of events as explained in~\Cref{subsec:preliminiaries-general}.} to be the \emph{\lpmtext{} occurrence list} of $\lpmvar$ in $\eventlogvar$.
    \end{definition}
    Additionally, we extend $\lpmocurrencelist$ with $\lpmocurrencelist_n$ such that the length of the minimal event \mbox{over-sequence} does not exceed $n$, i.e.,$\lpmocurrencelist_n(\lpmvar,\eventlogvar) = \set{\eventseqvar \st
    \exists_{\tracevar \in traces(\eventlogvar)} (\eventseqvar \relaxsub \tracevar \land \length{\overseq{\eventseqvar}{\tracevar}} \leq n
    \land \geteventattr{\eventseqvar}{\textit{act}} \in \lang{\lpmvar})}$

    Consider again the \lpmtext{} in~\Cref{fig:example-lpm-gp-log} and the event log extract given in~\Cref{tab:example-event-log-gameplay}.
    The \lpmtext{} occurrence list for $\lpmvar_{gp}$ and $\eventlogvar_{gp}$ is
    $\lpmocurrencelist(\lpmvar_{gp},\eventlogvar_{gp}) = \set{\seq{e_1, e_2, e_6}, \seq{e_4, e_5, e_6}}$ and
    $\lpmocurrencelist_4(\lpmvar_{gp}, \eventlogvar_{gp}) = \set{\seq{e4, e5, e6}}$ since
    $\length{\overseq{\seq{e1, e2, e6}}{\tracevar^{\eventlogvar_{gp}}_{202}}} = \length{\seq{e1, e2, e4, e5, e6}} = 5 > 4$.

    With the \lpmtext{} occurrence list, we can measure conformance to an event log $\eventlogvar$ and rank the \lpmtext{}s.
    The ranking can also take different quality measures into consideration, such as fitness, precision, and simplicity.
    We write $\rank \in \lpmuniverse \partmapsto \mathbb{N}$ to denote a ranking function, and $\rankingfuncuniverse$
    to denote the universe of all such ranking functions.


    \section{The LPM Grouping Framework}\label{sec:grouping-framework}
    In this part, we detail the proposed approach of grouping \lpmtext{}s, starting from a set of \lpmtext{}s and the
    event log on which they were discovered, all the way until the final \lpmtext{} groups and the chosen
    representatives for each group.

    \begin{figure}[t]
        \centering
        \includegraphics[width=\linewidth]{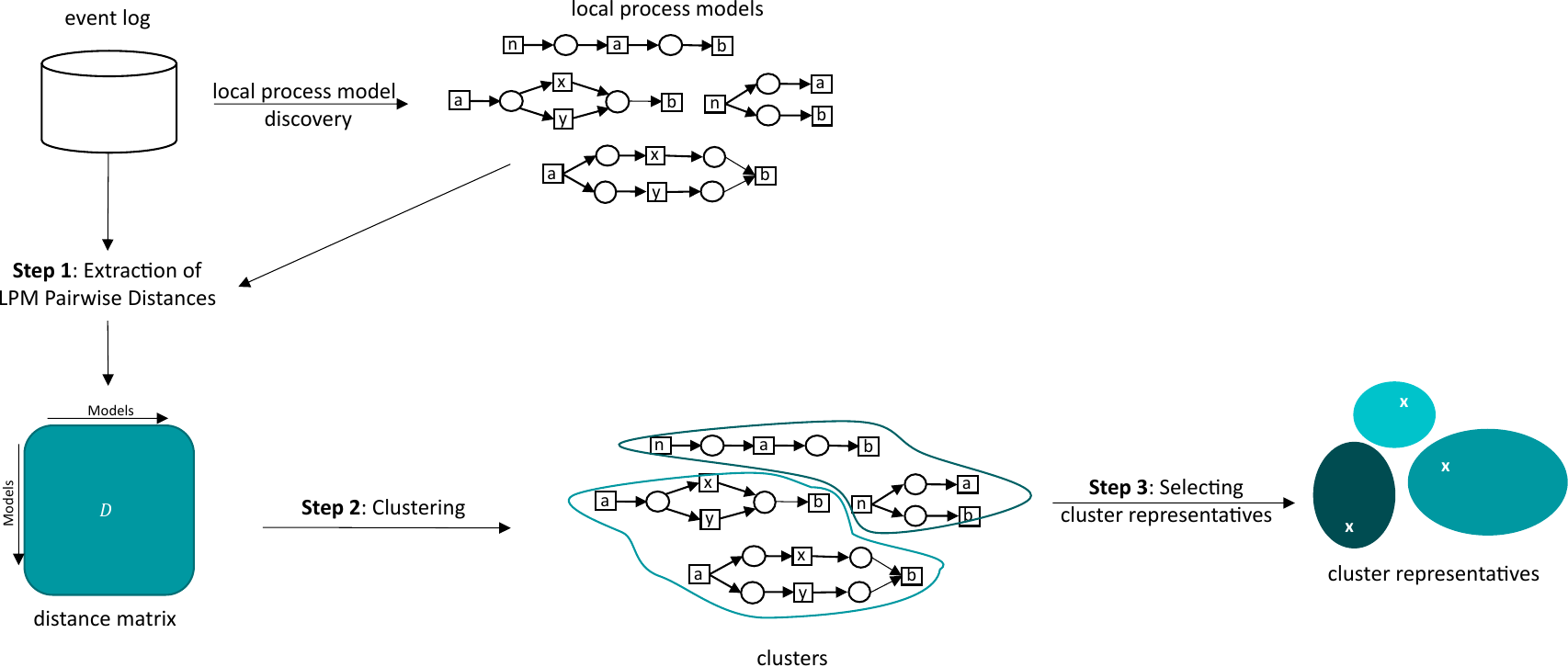}
        \caption{Illustration of the proposed three-step framework.}
        \label{fig:method}
    \end{figure}

    We give an overview of the grouping framework in~\Cref{fig:method}.
    The framework requires a set of \lpmtext{}s $\lpmsetvar$ and an event log $\eventlogvar$ as input.
    In the end, it returns a partitioning of the original \lpmtext{} set $\partition^{\lpmsetvar}$ and a set of representative \lpmtext{}s $\partition^{\lpmsetvar}_{\textit{repr}}$, i.e., a subset of \lpmtext{}s of the original set, each one belonging to exactly one group.
    We illustrate the entire approach by splitting it to individual steps.
    In the first step we compute the distance between two \lpmtext{}s using process model similarity measures, data attributes, and a combination of both.
    Next, we formally define clustering the set of \lpmtext{}s by using the precomputed distances.
    Finally, we choose one representative \lpmtext{} for each cluster and create the set of \lpmtext{} representatives.

    \subsection{Extraction of \lpmtext{} Pairwise Distances}\label{subsec:method-distance-matrix}
    In this step, we start with a set of \lpmtext{}s $\lpmsetvar$ and an event log $\eventlogvar$.
    For each \lpmtext{} pair we compute a distance between $0$ and $1$ using a distance function
    $\distfunc_x \in \distancefuncuniverse$ (where $\distancefuncuniverse \colon \lpmuniverse \times \lpmuniverse \to [0,1]$).
    The framework allows for three options for distance extraction: process model similarity measures, data attributes,
    and mixed.
    In the following subsection we explain the different distance extraction methods.

    \subsubsection{Process Model Similarity Measures}

    As discussed in \Cref{subsec:rw-grouping-process-models}, the literature agrees on two orthogonal
    categorizations of process model similarity measures.
    On the one hand, we have measures that focus on the model's structure or the model's behavior, or we include both.
    On the other hand, there are different abstraction levels at which a measure can consider the models, e.g.,
    complete language vs a directly-follow or an eventually-follow relationship abstraction.
    Here, we describe five measures that are representative of the abovementioned categories.
    We chose these five measures as a starting point for the grouping framework.
    However, the framework per design is not restrictive regarding the used process model similarity measures.

    To introduce the similarity measures, we assume we are given two \lpmtext{}s
    $\modelA = (N^{A}_{\lpmvar}, \emarking, \emarking)$ s.t. $\lpmnet{\lpmvar}^{A}=(P_A, T_A, F_A, l_A)$ and
    $\modelB = (\lpmnet{\lpmvar}^{B}, \emarking, \emarking)$ s.t. $N^{B}_{\lpmvar}=(P_B, T_B, F_B, l_B)$.
    In the following, we illustrate the measures we use in this work with the help of $\modelA$ and $\modelB$.

    \emph{Transition Label (TL) overlap} is the simplest measure we investigate.
    It computes the similarity between models by quantifying the TL overlap as shown below.
    \begin{equation*}
        \simm{\textit{transition}}(\modelA, \modelB) = \frac{2 * \length{l_A(T_A) \cap l_B(T_B)}}{\length{l_A(T_A)}+
        \length{l_B(T_B)}}
    \end{equation*}

    \emph{Node Matching (NM)} is somewhat more complex, in that it includes place overlap as well.
    We use this measure to represent structural measures using abstraction.
    The measure calculates the similarity between two models by combining TL overlap and place matching
    between the nets.
    We assign to each pair of places a matching gain $g(p_1, p_2) = \frac{1}{2}*\frac{2 * \length{l_A(\preset{p_1}) \cap
    l_B(\preset{p_2})}}{\length{l_A(\preset{p_1})} + \length{l_B(\preset{p_2})}} + \frac{1}{2}*\frac{2 * \length{l_A(
        \postset{p_1}) \cap l_B(\postset{p_2})}}{\length{l_A(\postset{p_1})} + \length{l_B(\postset{p_2})}}$, and we use
    the Hungarian algorithm~\cite{DBLP:books/daglib/p/Kuhn10} to solve the assignment problem.
    We use $G_{\textit{places}}$ to represent the gain of the optimal assignment.
    Then, we define the measure as
    \begin{equation*}
        \simm{\textit{node}}(\modelA, \modelB)=\frac{2 * \length{l_A(T_A) \cap l_B(T_B)} + 2 * G_{\textit{places}}}{\length{l_A(T_A)}+\length{l_B(T_B)} + \length{P_A} + \length{P_B}}
    \end{equation*}

    \emph{Eventually-Follow (EF) overlap} is a behavioral abstraction measure that measures the overlap of the
    eventually-follows relation in the languages of the two models.
    We calculate it as
    \begin{equation*}
        \simm{\textit{efg}}^n(\modelA, \modelB) = \frac{2 * |\textit{EF}^{\, n}_A \cap \textit{EF}^{\, n}_B|}{|\textit{EF
        }^{\, n}_A|+|\textit{EF}^{\, n}_B|}
    \end{equation*}
    such that $\textit{EF}_A^{\, n} = \set{(a, b) \st \exists_{\tracevar \in \nlang{\modelA}{n}}(\exists_
            {1 \leq i < j \leq \length{\tracevar}}(a = \tracevar(i) \land b = \tracevar(j)))}$ and $\textit{EF}^{\, n}_B$ is
    defined
    correspondingly.

    \emph{Trace Matching (TM)} is a similarity measure based on the behavior of the process models.
    Given the set of all traces of $\modelA$ and $\modelB$ restricted to a length at most $n$, i.e., $\nlang{\modelA}{n}$ and $\nlang{\modelB}{n}$, we assign each pair of traces $\tracevar_1 \in \nlang{\modelA}{n}$ and $\tracevar_2 \in \nlang{\modelB}{n}$ a matching gain $g(\tracevar_1, \tracevar_2)$.
    We compute the gain between two traces by inverting the normalized Levenshtein distance.
    Then, we use the Hungarian algorithm to solve the assignment problem.
    We use $G_{\textit{traces}}$ to represent the gain of the optimal trace assignment.
    At last, we define the measure as
    \begin{equation*}
        sim^n_{\textit{full}}(\modelA, \modelB) = \frac{2 * G_{\textit{traces}}}{\length{\nlang{\modelA}{n}} +
        \length{\nlang{\modelB}{n}}}
    \end{equation*}

    Finally, \emph{Graph Edit Distance (GED)} represents advanced structural measures.
    It calculates model similarity by computing the GED between two model.
    In our work, we use the measure as defined in~\cite[Definition~4]{DBLP:conf/bpm/DijkmanDG09}.

    Note that all similarity measures compute values between $0$ and $1$.
    In~\Cref{def:dist-lpms-pmsm} we formally define the conversion to a distance measure.

    \begin{definition}[\lpmtext{}s Distance For Process Model Similarity Measures]
        Given two \lpmtext{}s $\modelA$ and $\modelB$, and a similarity measure $\simm{}$, we compute the distance between them by inverting the computed similarity.
        Formally, $\distfunc(\modelA, \modelB) = 1 - \simm{}(\modelA, \modelB)$.
        \label{def:dist-lpms-pmsm}
    \end{definition}

    \subsubsection{LPM Data Attributes Measures}
    In this section, we use the events covered by an \lpmtext{} and the available data attributes to represent the context in which an \lpmtext{} appears in the process.
    To get the events covered by an \lpmtext{} $\lpmvar$, we use the occurrence list from~\Cref{def:lpm-occurrence-list}.
    In the continuation, we first describe encoding individual data attributes, followed by the process of appending the individual contexts to form the collective one.
    Then, we define how to compute the distance between two \lpmtext{}s by computing the distance between their collective context vectors.

    \paragraph{Context Encoding of Data Attributes}
    Given an event log $\eventlogvar = (\eventlogeventsvar, \eventattributes, \eventmapping)$, an \lpmtext{} $\lpmvar$,
    and a data attribute $\eventattributevar \in \eventattributes$, we obtain the multiset of events covered by $\lpmvar$ in $\eventlogvar$ using the occurrence list function, that is $\eventlogeventsvar^{\eventlogvar}_{\lpmvar} = \multiset{\eventvar \in \eventseqvar \st \eventseqvar \in \lpmocurrencelist(\lpmvar, \eventlogvar)}$ are the \emph{covered events}.
    Then, we compute the \emph{value multiset} for $\eventattributevar$ and $\lpmvar$ in $\eventlogvar$ by
    projecting the covered events on the attribute of interest $\eventattributevar$: ${\eventattributesvalues}^{(\eventlogvar, \lpmvar)}_{\eventattributevar} = {\geteventattr{\eventlogeventsvar^{\eventlogvar}_{\lpmvar}}{\eventattributevar}}$.
    Now, given the value multiset, we extract a data vector $\vec{d}^{(\lpmvar, \eventlogvar)}_{\eventattributevar}$ to represent the data attribute context for $\lpmvar$ in $\eventlogvar$.
    Depending on whether the data attribute is categorical or numerical we use different strategy to build the data
    vector, as explained below.

    \emph{Numerical Data Attributes.}
    Given the value multiset ${\eventattributesvalues}_{\eventattributevar_{n}}$ (short $V$) for a numerical data
    attribute $\eventattributevar_n$, we then extract four features to represent the data.
    \begin{itemize}
        \item The minimum value $\textit{min}(V) = \min\limits_{x \in V} x$
        \item The maximum value $\textit{max}(V) = \max\limits_{x \in V} x$
        \item The mean: $\textit{mean}(V) = \frac{\sum\limits_{x \in V}{x}}{\length{V}}$
        \item The median\footnotemark{}: $\textit{median}(V) =
        \begin{cases}
            \seqvar_V(\frac{\length{V} - 1}{2}), & \text{if } \length{V} \equiv 1 \mod 2 \\
            \frac{\seqvar_V(\frac{\length{V}}{2}) + \seqvar_V(\frac{\length{V} - 2}{2})}{2}, & \text{if } \length{V} \equiv 0 \mod 2
        \end{cases}$
    \end{itemize}
    \footnotetext{To demonstrate the median computation, we use the notation $\seqvar_M$ to represent the sequence
    computed from a multiset $M \in \allmultiset{\mathbb{N}}$ using the total order defined by the $\leq$ operator.}
    From these four features, we create the data vector that represents the numerical data attribute in the still to be
    built collective context for the \lpmtext{}.
    For example, consider the numerical attribute \constname{Idle Units} in $\eventlogvar_{gp}$ (\Cref{tab:example-event-log-gameplay}).
    The computed multiset for this data attribute and the \lpmtext{} $\lpmvar_{gp}$ in~\Cref{fig:example-lpm-gp-log}
    is
    $\eventattributesvalues^{(\lpmvar_{gp}, \eventlogvar_{gp})}_{\textit{idle-units}} = \geteventattr{\multiset{e1^1, e2^1, e4^1, e5^1, e6^2}}{\textit{idle-units}} = \multiset{27, 23, 31, 16^3}$.
    The resulting data vector consisting of the four features is
    $\begin{bmatrix}
         16 & 31 & \frac{\sum_{x \in \multiset{27, 23, 31, 16^3}}{x}}{6} & \frac{16+23}{2}
    \end{bmatrix}$.

    \emph{Categorical Data Attributes.}
    Given the value multiset ${\eventattributesvalues}_{\eventattributevar_{c}}$ for a categorical data attribute
    $\eventattributevar_c$, we create a feature vector where each distinct value
    represents a feature and the values are the count of that value appearing in the multiset.
    Consider the categorical attribute \constname{Experience Tier} in $\eventlogvar_{gp}$ (\Cref{tab:example-event-log-gameplay}).
    The computed multiset for this data attribute and the \lpmtext{} $\lpmvar_{gp}$ in~\Cref{fig:example-lpm-gp-log} is
    $\eventattributesvalues^{(\lpmvar_{gp}, \eventlogvar_{gp})}_{\textit{tier}} = \geteventattr{\multiset{e1^1, e2^1, e4^1, e5^1, e6^2}}{\textit{tier}} = \multiset{medium^6}$.
    We then convert the computed multiset to the data vector $\begin{bmatrix}
                                                                  0 & 6
    \end{bmatrix}$, where the first value corresponds to the data value \constname{advanced} and the second value to the
    data value \constname{medium}.

    \paragraph{Encoding of The Collective Occurrence Context}
    In this part, we use the context encoding of single data attributes to produce the collective occurrence
    context by first normalizing and then joining the single context encodings.
    Let $D^{(\lpmvar, \eventlogvar)}$ be the set of data vectors of all data attributes
    $\eventattributevar \in \eventattributes$ for one \lpmtext{}, and
    $D^{\eventlogvar} = \bigcup_{\lpmvar \in \lpmsetvar} D^{(\lpmvar, \eventlogvar)}$ the set of data vectors of
    all data attributes for all \lpmtext{}s in $\lpmsetvar$.

    To ensure that each data attribute contributes the same to the collective context, we normalize the computed data
    vectors $D^{(\lpmvar, \eventlogvar)}$ across all attributes.
    Since the data vectors for different data attributes vary in length, we normalize them by its length, i.e.,
    $\vec{d} = \frac{\vec{d}}{\norm{\vec{d}}}$ where $\vec{d} \in D^{\eventlogvar}$.
    This way, each data attribute plays an equal role in the collective context.

    Finally, in~\Cref{def:collective-context}, we define how we build the collective context for one \lpmtext{}, by first
    ordering the set of data vectors $D^{(\lpmvar, \eventlogvar)}$, so that each position always represents the same
    feature across all \lpmtext{}s, and concatenating them together.
    We use $(D^{(\lpmvar, \eventlogvar)}, \leq)$ to denote the ordered set.

    \begin{definition}[Collective Context]
        Given an event log $\eventlogvar$, an \lpmtext{} $\lpmvar$, a set of data attributes $\eventattributes$, the precomputed and normalized set of all data vectors $D^{(\lpmvar, \eventlogvar)}$ for those attributes, and an ordering $\leq_{\eventattributes}$, we call the data vector $\collectivecontext^{(\lpmvar, \eventlogvar)} = \concat (D^{(\lpmvar, \eventlogvar)}, \leq_{\eventattributes})$, formed as concatenation of all the data vectors, the \emph{collective context} of the \lpmtext{} $\lpmvar$ in $\eventlogvar$.
        \label{def:collective-context}
    \end{definition}

    \paragraph{Distance Computation Between Data Vectors}
    To compute the distance between two \lpmtext{}s, we employ distance functions that compute the distance between
    two data vectors representing their respective collective contexts.
    Well-known distance functions of this form are the Euclidean distance, Chebyshev distance, cosine distance, Minkowski distance, etc.
    We use the notation $\distfunc_v$ to represent such functions without explicitly selecting one.
    Most of these distances do not give a value between $0$ and $1$, so we additionally normalize the distance.

    Finally, in~\Cref{def:dist-lpms-data-attributes} we instantiate the distance computation between two \lpmtext{}s in~\Cref{def:dist-lpms-data-attributes} when we consider data attributes.

    \begin{definition}[\lpmtext{}s Distance For Data Attributes]
        Given an event log $\eventlogvar$, and two \lpmtext{}s $\modelA$ and $\modelB$, we compute the distance between $\modelA$ and $\modelB$ as the distance between the data vectors of their corresponding collective contexts $\collectivecontext^{(\modelA, \eventlogvar)}$ and $\collectivecontext^{(\modelB, \eventlogvar)}$.
        Formally, $\distfunc(\modelA, \modelB) = \distfunc_v(\collectivecontext^{(\modelA, \eventlogvar)},\collectivecontext^{(\modelB, \eventlogvar)})$, where $\distfunc_v$ computes the normalized distance between two data vectors.
        \label{def:dist-lpms-data-attributes}
    \end{definition}

    \subsubsection{Combined Measures}
    To let both similarities in the process model and in the context of which \lpmtext{}s appear weigh-in in building the groups, we allow for computing a mixed distance between two \lpmtext{}s, $\modelA$ and $\modelB$.
    To achieve this, we introduce a threshold $\tau \in [0, 1]$ that controls the influence of the distance computed using process model similarity measures $dist^{pms}$ versus the distance computed from the collective contexts $dist^{cc}$.
    We compute the mixed distance as $dist^{mx} = \tau * dist^{pms} + (1 - \tau) * dist^{cc}$.

    We also extend the notation of mixed distance, by allowing multiple distances computed from process model similarity measures or data vectors for specific subsets of attributes.
    Let $\tau_i \in [0, 1]$ for $i \in \set{1, ... k}$ such that $\sum_{i=1}^{k} \tau_i = 1$.
    We define the mixed distance as $dist^{mx} = \sum_{i=1}^{k} \tau_i * dist_i$ where $dist_i \in \set{dist^{pms}_x, dist^{cc}_x}$ and $x$ denotes the concrete process model similarity measure used or the subset of attributes to compute the data vector.

    \subsection{Clustering}\label{subsec:method-clustering}
    In the clustering step, we use a distance function $dist \in \distancefuncuniverse$ from the ones defined previously, to partition the set $\lpmsetvar$ in groups.
    To formalize this, we define a \emph{clustering algorithm} in \Cref{def:clustering}.

    \begin{definition}[Clustering Algorithm]
        \label{def:clustering}
        Let $\clustersetuniverse = \set{X \subseteq \powerset{\lpmuniverse}}$ be the universe of cluster sets.
        A function $\textit{clust} \in \powerset{\lpmuniverse} \times \distancefuncuniverse \partmapsto \clustersetuniverse$ is a \emph{clustering algorithm} if and only if for any \lpmtext{} set $\lpmsetvar \in \powerset{\lpmuniverse}$ and a distance function $\distfunc \in \distancefuncuniverse$, $X_{\lpmsetvar} = \textit{clust}(\lpmsetvar, \distfunc)$ satisfies the following two properties:
        \begin{itemize}
            \item $\bigcup X_{\lpmsetvar} = \lpmsetvar$
            \item $\forall_{X_i, X_j \in X_{\lpmsetvar}} X_i \cap X_j = \emptyset$
        \end{itemize}
        To denote a clustering algorithm given some set of parameters $P$, we write $\textit{clust}_P$.
    \end{definition}

    The goal of the clustering algorithm is to return the \lpmtext{}s in homogeneous groups, such that the similarity is high within the individual groups and low between them.
    One clustering algorithm can produce different cluster sets for the same distance function and set of \lpmtext{}s based on the parameters $P$.

    Now, given the predefined distance function $\distfunc$, an \lpmtext{} set $\lpmsetvar$, and a \emph{clustering algorithm} $\textit{clust}_P$ we compute a set of clusters $\partition^{(\mathit{clust}_P, \lpmsetvar, \distfunc)} = \mathit{clust}_P(\lpmsetvar, \distfunc) = X_{\lpmsetvar}$ such that $X_{\lpmsetvar} \in \clustersetuniverse$.
    We overload the notation $\partition^{(\mathit{clust}_P, \lpmsetvar, \distfunc)}(\lpmvar)$ to denote the cluster $X \in X_{\lpmsetvar}$ for which the \lpmtext{} $\lpmvar$ belongs, i.e., $\lpmvar \in X$.
    That is, it holds $\lpmvar \in \partition^{(\mathit{clust}_P, \lpmsetvar, \distfunc)}(\lpmvar) \in \partition^{(\mathit{clust}_P, \lpmsetvar, \distfunc)}$.

    \subsection{Computing Cluster Representatives}\label{subsec:method-cluster-representatives}
    In Step 3, we use the computed cluster set $\partition^{(\mathit{clust}_P, \lpmsetvar, \distfunc)}$ from Step 2, to calculate the \emph{representative set} $\partition^{(\mathit{clust}_P, \lpmsetvar, \distfunc)}_{\textit{repr}} \subseteq \lpmsetvar$ in which we keep only one representative \lpmtext{} per cluster.
    In \Cref{def:repr}, we formally define \emph{representative projection} as a function that maps a set of \lpmtext{}s to one model.

    \begin{definition}[\lpmtext{} Set Representative]
        \label{def:repr}
        Given a ranking function $\rank \in \rankingfuncuniverse$ and a set of \lpmtext{}s $\lpmsetvar$, we define $\textit{repr}(\lpmsetvar, \rank) = \argmax_{\lpmvar \in \lpmsetvar} \rank(\lpmvar)$ to be the \emph{representative} of $\lpmsetvar$ given $\rank$.
    \end{definition}

    Now, for the set of \lpmtext{}s $\lpmsetvar \subseteq \lpmuniverse$ and a ranking function $\rank$, we create the set
    $\partition^{(\mathit{clust}_P, \lpmsetvar, \distfunc)}_{\textit{repr}} = \set{\textit{repr}(\lpmsetvar_i, \rank)
        \st \lpmsetvar_i = \partition^{(\mathit{clust}_P, \lpmsetvar, \distfunc)}(\lpmvar) \land \lpmvar \in \lpmsetvar}$,
    and we call it the \emph{cluster representatives}.
    This way, we significantly reduce the number of \lpmtext{}s from the original set $\lpmsetvar$, but still keep the
    essence of the entire set.
    As a ranking function we use $\rank^{\textit{freq}}_{\eventlogvar}(\lpmvar) = \length{\lpmocurrencelist(\lpmvar, \eventlogvar)}$
    to take the most frequent model, or
    $\rank^{\textit{dist}}(\lpmvar) = 1 - \frac{\sum_{\lpmvar' \in \partition(\lpmvar)} \distfunc(\lpmvar, \lpmvar')}{\length{\partition(\lpmvar)}}$
    to take the medoid \lpmtext{}, i.e., the model with the minimal distance to all other models in the cluster.

    \subsection{Conclusion}\label{subsec:method-conclusion}
    To summarize, in our grouping framework we start with an event log $\eventlogvar$ and a set of \lpmtext{}s $\lpmsetvar$.
    The framework allows for computing the distance between \lpmtext{}s using established process model similarity measures or by representing each \lpmtext{} as a data vector summarizing data attributes of interest for the covered events.
    Given we chose one of the ways to compute the distance, we use a clustering algorithm to group similar \lpmtext{}s together.
    Finally, given a ranking function $\rank$, for each group we extract one \lpmtext{} that serves as a representative of that group.
    The final outputs of the framework are the set of clusters and the cluster representatives.


    \section{Evaluation of the Framework}\label{sec:evaluation_results}

    To evaluate the proposed framework, we consider several perspectives.
    First, to assess the impact of the framework, we compare the top-scoring samples of \lpmtext{}s with
    representative samples (for different sample sizes).
    Second, we evaluate the cohesiveness of the clustering results and if those correlate with the repetitiveness and
    coverage computed before.
    Third, we report the running time of the distance computation and clustering relative to the discovery approach.
    Finally, we examine three representative \lpmtext{}s for the event log used as motivational example in
    \Cref{sec:introduction}.

    We organize the rest of the section by first presenting implementation and the experimental setup.
    Then, we have a section for each of the perspectives we evaluate.

    \subsection{Implementation}
    We implemented the entire framework in \texttt{ProM}, as part of the
    \texttt{LocalProcessModelDiscoveryByCombiningPlaces} package\footnote{\url{https://github.com/promworkbench/LocalProcessModelDiscoveryByCombiningPlaces/releases/tag/jiis24-grouping-lpms}}.
    The same package also contains the \lpmtext{} discovery approach presented in~\cite{DBLP:conf/apn/PeevaMA22}.

    In the implementation, we chose hierarchical clustering algorithm for grouping the \lpmtext{}s and considered
    number of clusters and linkage as possible parameters.
    Linkage determines how the distance between two clusters containing multiple models is calculated.
    For the hierarchical clustering algorithm and everything around it, we use the Smile library in Java~\cite{Li2014Smile}.
    All other computations for the process model similarity measures and encoding the context using the data
    attributes are directly implemented in the package itself.

    \subsection{Experimental Setup}
    We perform all experiments on the event logs and \lpmtext{} sets in~\Cref{tab:sets}.
    For computing distances, we use the five process model similarity measures we described and the data attribute
    measure.
    For the hierarchical clustering, we change the linkage between: single, complete, upgma, and wpgma\footnote{For
    more information on the upgma and wpgma linkage, see \url{https://haifengl.github.io/clustering.html}.};
    and iterate the number of clusters for the following values: $\set{2, 5, 10, 15, 20, 30, 40, 50, 100, 200}$.
    Considering all combinations of an \lpmtext{} set, distance extraction method, and clustering algorithm parameter, we
    have $1440$ experiments in total.
    Due to space limitations, we only show the results of some of the experiments or a summary, but we upload all 
    obtained data for the experiments to \constname{Zenodo}\footnote{\url{https://zenodo.org/records/17013367}} and the
    analysis notebooks to
    \constname{Gitlab}\footnote{\url{https://git.rwth-aachen.de/evaluations/jiis2024-groupinglpms}} for interested
    readers.

    \begin{table}[t]
        \centering
        \caption{Local process model sets used in the evaluation}
        \begin{tabular}{c|c|c}
            \hline
            \textbf{Event Log} & \textbf{LPM set} & \textbf{Number of models} \\ \hline
            BPI Challenge 2012                        & $\lpmsetvar_{\textit{BPIC2012}}$          & $547$ \\ \hline
            BPI Challenge 2012 - resource 10939       & $\lpmsetvar_{\textit{BPIC2012-res10939}}$ & $562$ \\ \hline
            BPI Challenge 2017 - Offers Log           & $\lpmsetvar_{\textit{BPIC2017}}$          & $578$  \\\hline
            BPI Challenge 2020 - Prepaid Travel Costs & $\lpmsetvar_{\textit{BPIC2020}}$          & $292$  \\ \hline
            Sepsis                                    & $\lpmsetvar_{\textit{Sepsis}}$            & $505$  \\ \hline
            Hospital Billing                          & $\lpmsetvar_{\textit{HB}}$                & $626$ \\ \hline
        \end{tabular}
        \label{tab:sets}
    \end{table}

    \subsection{\lpmtext{} Samples Comparison}\label{subsec:evaluation-lpm-diversity-analysis}
    The focus of the paper is to showcase that a sample of representative \lpmtext{}s are better suited to capture
    the underlying process than a sample of top-scoring \lpmtext{}s.
    Since the capturing of the process' essence is an abstract utility, we quantify it using two measures that serve as
    proxies:
    \begin{enumerate}
        \item \emph{Coverage} - the fraction of events covered by the full set of \lpmtext{}s that are also covered by the
        sample.
        Given an event log $\eventlogvar$, an \lpmtext{} set $\lpmsetvar$, and a
        sample $\lpmsetvar_s \subseteq \lpmsetvar$, we compute coverage as
        $cov = \frac{\length{\set{e \in s \st s \in \lpmocurrencelist(\lpmvar,\eventlogvar) \land \lpmvar \in \lpmsetvar_s}}}
        {\length{\set{e \in s \st s \in \lpmocurrencelist(\lpmvar,\eventlogvar) \land \lpmvar \in \lpmsetvar}}}$.
        \item \emph{Diversity} - the mean pairwise distance between the \lpmtext{}s in the sample to quantify model
        repetition.
        Given an \lpmtext{} sample $\lpmsetvar$ and a distance function $\distfunc$, we compute diversity as
        $div = \frac{\sum_{\lpmvar_1, \lpmvar_2 \in \lpmsetvar, \lpmvar_1 \neq \lpmvar_2} \distfunc(\lpmvar_1, \lpmvar_2)}
        {\length{\lpmsetvar} \cdot (\length{\lpmsetvar} - 1)}$.
    \end{enumerate}
    Hence, this part of the evaluation compares the coverage and diversity of the top-scoring sample
    of \lpmtext{}s to the computed representative samples based on distance and score, for each clustering.

    \textbf{Setup.} First, for each clustering result, we derive three samples: (1) sample containing the highest
    scoring \lpmtext{}s in each of the resulting groups according to the $\rank_{\eventlogvar}^{\textit{freq}}$
    ranking (\emph{score representative sample} (RS)), (2) sample containing all the medoid \lpmtext{}s of the
    resulting groups, or the highest scoring \lpmtext{}s in each group according to the
    $\rank^{\textit{dist}}$ ranking (\emph{distance representative sample} (RD)), and (3) sample
    containing the top $k$ highest scoring \lpmtext{}s overall according to the $\rank_{\eventlogvar}^{\textit{freq}}$,
    where $k$ corresponds to the number of groups in the clustering (\emph{top sample} (T)).
    Then, for each sample, we compute \emph{coverage} and \emph{diversity} and compare them pairwise.

    Choosing any of the samples is deciding between the trade-offs associated with them.
    The T sample is expected to have high coverage, as it contains the most frequent \lpmtext{}s, but as
    illustrated in the example in \Cref{sec:introduction}, some of its coverage may be duplicated.
    Moreover, since the sample is selected without considering similarity between the \lpmtext{}s, it is expected to
    have low diversity.
    The RS sample is also expected to achieve high coverage, as it includes the most frequent \lpmtext{}s in each group.
    Since the groups are formed based on similarity, the RS sample is also expected to have higher diversity than
    the T sample and mitigate duplicate coverage.
    The RD sample, including the medoid of each group, is expected to have high diversity.
    However, since the medoids are not selected based on frequency, the coverage of the RD sample is sacrificed twice:
    first, by grouping similar \lpmtext{}s together, and second, by not selecting the most frequent \lpmtext{} in
    each group.
    We expect that, in terms of coverage, the RS and T samples will always be better than RD, and will alternate
    between each other depending on an event log, the clustering parameters, and similarity measures.
    In terms of diversity, we expect the order to be RD, RS, and last T.

    \begin{figure}[ht]
        \centering
        \includegraphics[width=\linewidth]{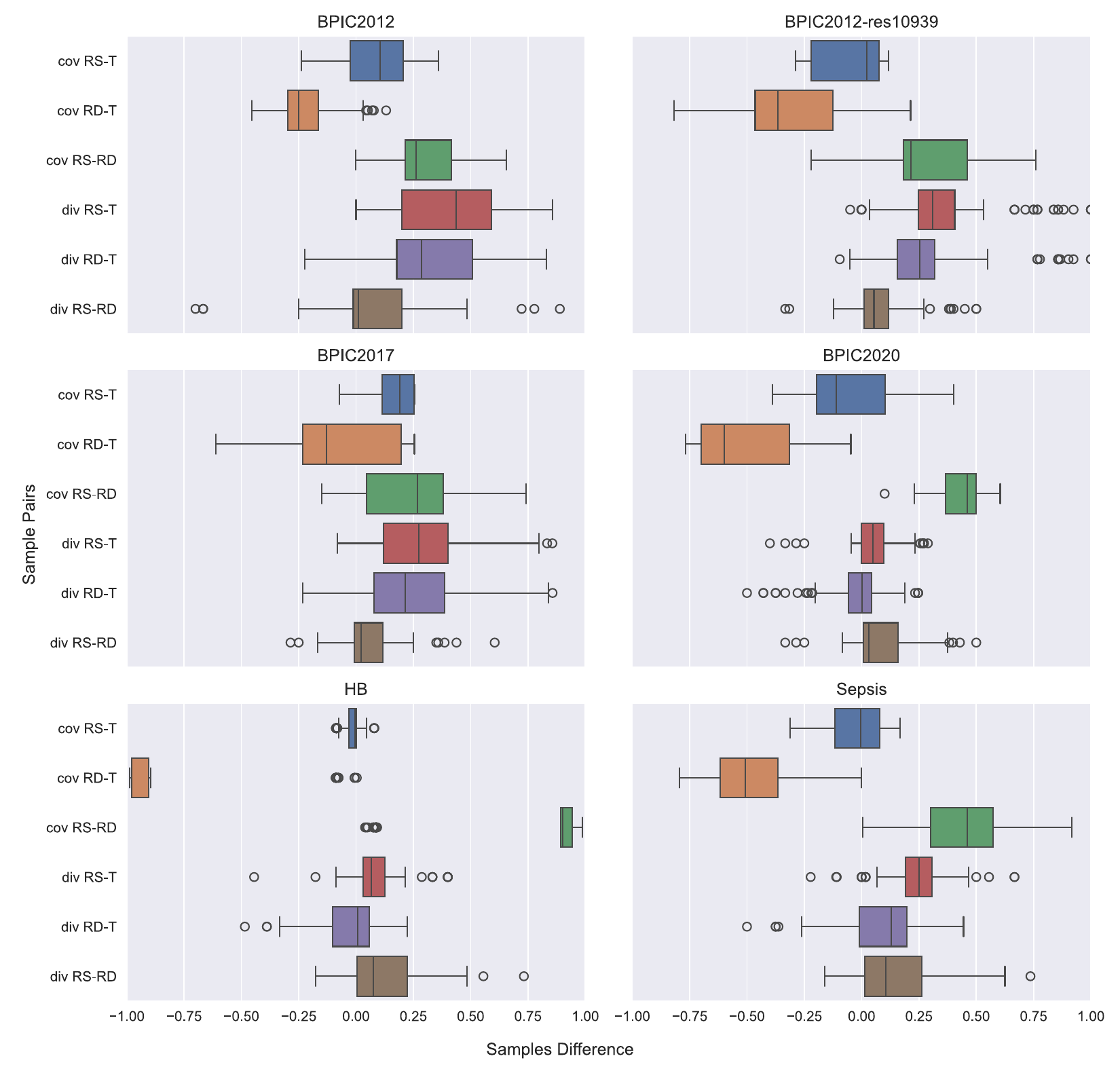}
        \caption{The difference in coverage and diversity between the samples. The notation \textit{cov S1-S2}
        represents values obtained using $cov_{S1} - cov_{S2}$, where $cov$ stands for coverage, $div$ for diversity,
        and $S1$, $S2$ are placeholders for T (top sample), RS (score representative sample), and RD (distance
        representative sample).}
        \label{fig:set_comparison}
    \end{figure}

    \textbf{Results.} To compare the samples, we compute the difference between the computed coverage and
    diversity values for each pair of samples per obtained clustering.
    Considering that the bigger the sample, the top-scoring sample would have an advantage, we restrict to
    $k \in \set{2, 5, 10, 15, 20}$.
    The results are shown for each event log separately on \Cref{fig:set_comparison}, where the x-axis represents the
    computed difference values and the y-axis the different pairwise sample comparisons for both measures (in total six).
    From the figure, we can make the following conclusions.

    The coverage computed for the RD sample is consistently lower than the coverage computed for
    the T and RS samples.
    This is visible from the orange and green boxplots.
    The orange boxplot represents $cov_{RD} - cov_{T}$ and shows that most values are negative.
    Thus, the coverage computed for the RD sample is in most cases lower than the coverage computed for the T sample.
    In contrast, the green boxplot represents $cov_{RS} - cov_{RD}$ and indicates that the majority of the observations
    fall into the positive range.
    Thus, the coverage computed for the RS sample is in most cases higher than the coverage computed for the RD sample.
    The coverage difference between T and RS samples, represented as a blue boxplot, varies between event logs.
    For the \constname{BPIC} event logs, in most cases the RS sample has higher coverage, while for \constname{HB},
    \constname{Sepsis}, and \constname{BPIC2020} event logs it is almost half-half split between negative and positive values.
    All results surrounding coverage support our initial expectations.

    The diversity is higher for almost all event logs for the RS and RD samples compared to the T sample (red and purple boxplots, respectively).
    Exceptions are the \constname{HB} and \constname{BPIC2020} event logs for the RD sample, where the values are
    equally distributed below and above 0, which is quite surprising.
    What is also surprising is that the RS samples score better on diversity than the RD samples (as indicated by the
    brown boxplots).
    A potential explanation would be that the representatives in the RD sample try to balance the distance to
    all \lpmtext{}s in the group, even those that also closely relate to \lpmtext{}s from other groups.
    Hence, the medoids in each group sometimes also overlap with the medoids in the complete set, resulting in
    medoids that are close in distance.
    Our assumption about the RD sample having the highest diversity does not hold, making the RD sample inferior to
    the RS sample in both diversity and coverage.

    \subsection{Correlating Cohesiveness to Representativeness}
    \label{subsec:evaluation-comparison-between-distance-methods}

    In the previous section, we found that RS samples outperform RD samples.
    Considering that the clustering results can influence the samples, we investigate whether these results and
    improvements compared to T samples correlate with cluster cohesiveness.
    We assess the cohesiveness of the clustering results by computing the silhouette score~\cite{ROUSSEEUW198753}.
    Most of the computed silhouette score values lie between $-0.5$ and $0.5$ indicating overlapping clusters for
    most distance measures, clustering parameters, and \lpmtext{} sets.
    The majority of the values above $0.5$ occur for the TL and DV measures, while the extremely negative values
    occur for NM and single linkage.
    We could not clearly identify any other patterns.

    \textbf{Setup.} We consider three subsets of the data for which we analyze correlation: (1) all $1440$
    clustering, (2) clustering where number of clusters is not larger than $20$, and (3) clustering where where
    number of clusters is not larger than $20$ and $div_{RS} - div_{RD} > 0$.
    For each subset, we compute the Pearson correlation coefficient between the silhouette score values and the
    difference values in coverage and diversity between the three samples we computed in
    \Cref{subsec:evaluation-lpm-diversity-analysis}.

    \textbf{Results.} None of the $18$ obtained values show an indication of a strong correlation between
    cohesiveness and representativeness.
    However, for $cov_{RS} - cov_T$ and $cov_{RD} - cov_T$, and subsets (2) and (3) introduced above, the values
    are around $0.3$, hinting a weak correlation.
    A significant jump in correlation is noticeable for the $div_{RS} - div_T$ and $div_{RD} - div_T$ values, from
    around $0$ when all clustering are considered to almost $0.3$ for clustering with not more than $20$ clusters.
    This jump and the improvement from $0.2$ to $0.3$ for the coverage, shows that our decision to restrict to
    clustering with not more than $20$ clusters in the previous evaluation was reasonable.
    All results obtained for clustering where $div_{RS} - div_{RD} > 0$ were very close to the results obtained for
    the subset containing clustering with no more than $20$ clusters.

    \subsection{Execution Time Analysis}\label{subsec:execution-time-analysis}
    Here, we investigate the time complexity of grouping the \lpmtext{}s and also relate it to the time needed for
    discovery.

    The execution time of the grouping framework depends mainly on two factors: (1) the distance computation
    between \lpmtext{}s, and (2) the clustering.
    For the distance computation, $20$ of the $36$ computations are below $10$ seconds, most of the rest are up to $20$
    seconds, one is $30$ seconds, and one is $100$ seconds.
    All above $10$ seconds are measures requiring generating the language of the models, TM and EF, and the GED measure.
    The highest computation is for the DV measure and the \constname{BPIC2017} event log.
    The clustering times are all below $3$ seconds.
    When comparing to the times needed for discovering the \lpmtext{}s, in the majority of experiments (around $1100$),
    the grouping including the distance computation between \lpmtext{}s, takes less than half the time the discovery
    needs.
    However, for a smaller subset (around $150$), the grouping can take up to three times the discovery time.
    It is worth noting that the reported times may vary from execution to execution.
    Therefore, to obtain more stable results, we need to let the experiments run a certain number of times and then
    produce aggregated values.

    \subsection{Case Study: BPIC2012-res10939}\label{subsec:evaluation-case-studies}

    \begin{figure}[t]
        \centering
        \begin{subfigure}[b]{0.6\textwidth}
            \includegraphics[width=\linewidth]{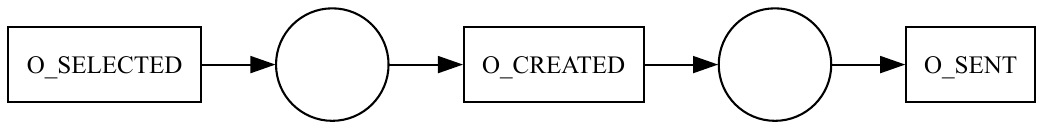}
            \caption{\lpmtext{} $1$ - original ranking $1$}
        \end{subfigure}
        \begin{subfigure}[b]{0.4\textwidth}
            \includegraphics[width=\linewidth]{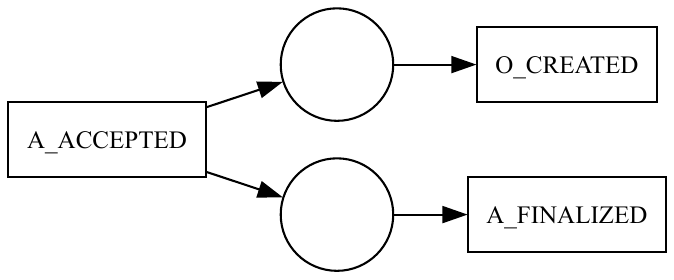}
            \caption{\lpmtext{} 2 - original rank $5$}
        \end{subfigure}
        \begin{subfigure}[b]{0.4\textwidth}
            \includegraphics[width=\linewidth]{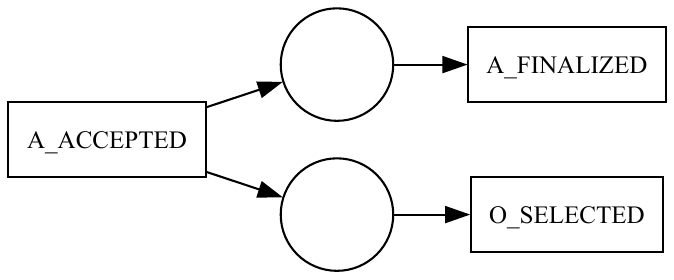}
            \caption{\lpmtext{} $3$ - original rank $7$}
        \end{subfigure}
        \caption{The three top-scoring representative \lpmtext{}s for \constname{BPIC2012-res10939}.}
        \label{fig:highest_ranked_repr_models}
    \end{figure}

    To complement the quantitative analysis above, we finish with examining three representative
    models for the \constname{BPIC2012-res10939} event log.
    The advantage of such a small case study is that it demonstrates the benefits of grouping from the user's
    perspective.

    In \Cref{fig:similar_models}, we showed that the top-scoring models for both~\cite{DBLP:journals/jides/TaxSHA16}
    and~\cite{DBLP:conf/apn/PeevaMA22} focus on different behavioral
    variants of \constname{O\_SELECTED}, \constname{O\_CREATED}, and \constname{O\_SENT}.
    Such repetition appears for lower-ranked models as well.
    In \Cref{fig:highest_ranked_repr_models}, we show the three top-scoring representative \lpmtext{}s after
    grouping the original set of \lpmtext{}s.
    To obtain the \lpmtext{}s, we use the results for the EF measure with wpgma linkage and $20$ clusters.
    It is clear that the collective language of these three models is larger than of those described by the three
    top-scoring original models discovered by both Peeva et al. and Tax et al. (see \Cref{fig:similar_models}).
    Additionally, the distance between the three representative models is $0.92$ while between the three
    top-scoring models is $0.18$.
    The distance between all representatives is also $0.92$ while between the $20$ top-scoring models is $0.58$.
    Moreover, the three top-scoring models cover $372$ out of the $2763$ events in the \constname{BPIC2012-res10939},
    while the three representative models cover $578$ events.
    Even more interesting is that the $20$ top-scoring models together cover also $578$ events or the same as
    the top three representatives, while all representatives together cover $658$ events.
    This combination is one example where grouping and selecting representatives improves both dimensions.


    \section{Discussion and Future Work}\label{sec:discussion}

    \subsection{Limitations}
    The \lpmtext{} grouping framework we presented, in theory, can be as flexible as possible regarding the
    representation of the allowed models, the way distance is measured, and clustering options.
    However, by offering a specific implementation of it, we made design decisions that restrict its potential and
    introduced different biases.

    \textbf{\lpmtext{} Representation.}
    In our framework, we restrict to \lpmtext{}s represented as accepting Petri nets.
    However, \lpmtext{}s can be depicted using any process modeling language as long as they represent patterns found in the data.
    Therefore, using a more different representation will also require adapting the distance measures.
    However, this is a limitation of the implementation, not of the framework.

    \textbf{Distance Measures.}
    We offer two ways to measure the distance between \lpmtext{}s: (1) structure and/or
    behavior and (2) the context in which they appear.
    For the process model similarity measures, we considered existing surveys and chose similarity
    measures representing the different categories in which the measures are divided.
    As part of future work, we can extend the framework with additional measures.

    For groping \lpmtext{}s based on the context in which they appear, we encode the context by using the event
    attribute values of events covered by an \lpmtext{}.
    Specifically, we summarize the values for each attribute using several descriptive indicators depending on the
    attribute type.
    By limiting to a specific set of indicators, we introduce bias in how we represent the context and when
    two \lpmtext{}s would be considered similar based on it.
    In future work, we can extend the offered descriptive indicators.
    However, the data attribute measure might be unusable for large-scale event logs with many attributes because
    of performance problems.
    Encoding categorical attributes using frequency vectors can result in sparse vectors, which in turn affects the
    complexity and robustness of the clustering.
    As an extension, the framework can support various heuristics for choosing valuable attributes and grouping infrequent categorical values.
    Furthermore, some activities can have more attributes than others, creating an imbalance in the context
    representation of \lpmtext{}s that include these activities.
    More specifically, the context and, thereby, the created groups will more heavily depend on the context of one
    activity (the one with more attributes) compared to the rest of the activities in the \lpmtext{}.
    Possible solutions would be to exclude non-overlapping attributes or use distance measures better suited for
    such situations.

    \textbf{Clustering.}
    We use hierarchical clustering as it does not require preselecting the number of clusters, is not dependent on
    many parameters, is more flexible regarding distance measures compared to other algorithms, and, depending on
    the linkage, it can find both spherical and non-spherical clusters.
    At the same time, hierarchical clustering is computationally heavy and might not be best suited for larger
    sets of \lpmtext{}s.
    As part of future work, we can extend the framework with additional clustering algorithms and parameters.

    \subsection{Evaluation Results}
    The evaluation we performed, while extensive, it is not comprehensive.
    \lpmtext{} discovery algorithms discover hundreds or thousands of models for one event log.
    Hence, we can assume that the variation of possible patterns explodes further when we consider processes in
    different contexts, different discovery algorithms, and parameters.
    Hence, the experiments we performed can be extended as part of future work.

    From our experiments, there are several conclusions we can make: (1) the clusters we get for the different
    event logs, distance measures, and parameters are mainly not cohesive, (2) despite obtaining non-cohesive
    clusters, the score representative sample computed from the grouping mostly scores better on the diversity measure
    and the coverage measure compared to when for each group the top-scoring \lpmtext{} is chosen, (3) cohesiveness
    does not correlate with representativeness, and (4) a few distance computations are execution heavy.
    Considering this, there are several considerations for the future.
    First, we can perform a root cause analysis on the results to uncover the source of the poor cluster compactness.
    Second, we should extend the framework with additional measures than the diversity and coverage.
    With the current results, there was no indication that more cohesive clusters produce more
    representative samples.
    However, there is no sufficient data with high-quality clustering results to be confident in the obtained results.
    Therefore, can be explored further in future work.


    \section{Conclusion}\label{sec:conclusion}
    In this paper, we described the challenges of model explosion and repetition in \lpmtext{} discovery, and
    motivated and proposed a framework for grouping similar \lpmtext{}s.
    We have two strategies to measure distance between two \lpmtext{}s.
    On the one hand, we use established process model similarity measures from the literature.
    On the other hand, we consider the events an \lpmtext{} covers, and use the data attributes available for those
    events to build an occurrence context of the \lpmtext{}.
    Then, we compute the distance between two \lpmtext{}s as the distance of the respective occurrence contexts.
    The proposed framework accepts an event log and a set of \lpmtext{}s as input, and consists of three-steps.
    In the first one, we use the aforementioned distance notions to compute pairwise distances between the \lpmtext{}s in the set.
    Then, the computed distances are used to cluster the set of \lpmtext{}s, and for each cluster, one \lpmtext{} is chosen as a cluster representative.
    In the evaluation, we showed how grouping similar \lpmtext{}s together improves process understandability on a real-life case study.
    Finally, we showcased \lpmtext{} diversity improvement on six real event logs for the original set of \lpmtext{}s
    and the computed set of representative \lpmtext{}s.
    We finished by discussing limitations of the implementation of the framework and the performed evaluation.

    \section*{Declarations}
    \subsection*{Ethics Approval}
    Not Applicable.
    \subsection*{Availability of supporting data}
    Not Applicable.
    \subsection*{Competing Interests}
    The authors declare no competing interests.
    \subsection*{Funding}
    Alexander von Humboldt-Stiftung and Bundesministerium für Bildung und Forschung, 16DHBKI016.
    \subsection*{Authors' Contributions}
    V.P. wrote the manuscript text and W.M.P v. d. A. reviewed the manuscript.
    \subsection*{Acknowledgments}
    \begin{minipage}[c]{0.3\textwidth}
        \includegraphics[width=\linewidth]{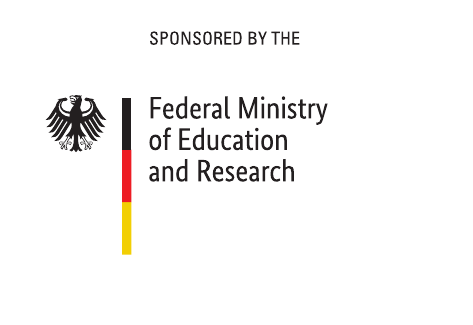}
    \end{minipage}\hfill
    \begin{minipage}[c]{0.7\textwidth}
        We thank the Alexander von Humboldt (AvH) Stiftung for supporting our research.
        The authors gratefully acknowledge the financial support by the Federal Ministry of Education and Research (BMBF) for the joint project AIStudyBuddy (grant no. 16DHBKI016).
    \end{minipage}

    \bibliography{bibliography}

\end{document}